\def\year{2021}\relax
\documentclass[letterpaper]{article} 
\usepackage{aaai21}  
\usepackage{times}  
\usepackage{helvet} 
\usepackage{courier}  
\usepackage[hyphens]{url}  
\usepackage{graphicx} 
\usepackage{natbib}  
\usepackage{caption} 
\usepackage{amssymb}
\usepackage{amsmath}
\usepackage{subfigure}
\usepackage{booktabs} 
\usepackage{mathtools}
\usepackage{multirow}
\usepackage{adjustbox}
\usepackage{xcolor}

\title{Randomly Weighted, Untrained Neural Tensor Networks Achieve Greater Relational Expressiveness}
\author{

    Jinyung Hong\textsuperscript{\rm 1}, Theodore P.~Pavlic\textsuperscript{\rm 1,2,3}
    \\
}
\affiliations{
	\textsuperscript{\rm 1} School of Computing,
	Informatics, and Decision Systems Engineering,
	Arizona State University, Tempe, AZ 85281, USA \\
	\textsuperscript{\rm 2} School of Sustainability,
	Arizona State University, Tempe, AZ 85281, USA \\
	\textsuperscript{\rm 3} School of Life Sciences,
	Arizona State University, Tempe, AZ 85281, USA \\
	\{jhong53, tpavlic\}@asu.edu

}

\begin{document}
\maketitle

\begin{abstract}
Neural Tensor Networks~(NTNs), which are structured to encode the degree
of relationship among pairs of entities, are used in Logic Tensor
Networks~(LTNs) to facilitate Statistical Relational Learning~(SRL) in
first-order logic. In this paper, we propose Randomly Weighted Tensor
Networks~(RWTNs), which incorporate randomly drawn, untrained tensors
into an NTN encoder network with a trained decoder network.
We show that RWTNs meet or surpass the performance of traditionally
trained LTNs for Semantic Image Interpretation~(SII) tasks that have
been used as a representative example of how LTNs utilize reasoning over
first-order logic to exceed the performance of solely data-driven
methods. We demonstrate that RWTNs outperform LTNs for the detection of
the relevant \emph{part-of} relations between objects, and we show that
RWTNs can achieve similar performance as LTNs for object classification
while using fewer parameters for learning. Furthermore, we demonstrate
that because the randomized weights do not depend on the data, several
decoder networks can share a single NTN, giving RWTNs a unique economy
of spatial scale for simultaneous classification tasks.
\end{abstract}

\section{Introduction}
\label{sec:intro}

Combining knowledge-representation-and-reasoning techniques with
artificial neural networks has the promise of enhancing the high
performance of modern artificial intelligence~(AI) with explainability
and interpretability, which are necessary for generalized human insight
and increased trustworthiness. Several recent studies across statistical
relational learning~(SRL), neural-symbolic computing, knowledge
completion, and approximate inference~\citep{koller2007introduction,
garcez2008neural, pearl2014probabilistic, nickel2015review} have shown
that neural networks can be integrated with logical systems to achieve
robust learning and efficient inference as well as the interpretability
provided by symbolic knowledge extraction.

These approaches to neural-network knowledge representation make use of
\emph{relational embedding}, which represents relational predicates in a
neural network~\citep{sutskever2009using, bordes2011learning,
socher2013reasoning, santoro2017simple}. For example, Neural Tensor
Networks~(NTNs) are structured to encode the degree of relationship
among pairs of entities in the form of tensor operations on real-valued
vectors~\citep{socher2013reasoning}. These NTNs have been synthesized
with neural symbolic integration~\citep{garcez2008neural} in the
development of Logic Tensor Networks~(LTNs)~\citep{serafini2016logic},
which are able to extend the power of NTNs to reason over first-order
many-valued logic~\citep{bergmann2008introduction}.

In this paper, we propose Randomly Weighted Tensor Networks~(RWTNs), a
novel NTN-based network for relational embedding that incorporates
randomly drawn, untrained tensors as an encoder network with a trained
decoder network. Our approach is motivated by the basic architecture of
an LTN combined with insights from Reservoir
Computing~(RC)~\citep{jaeger2001echo}, which is more traditionally
applied to classification problems and time-series analysis. A
conventional LTN would incorporate an NTN specially trained to capture
logical relationships present in data. In our case, the NTN we use is
selected not by training but by drawing a 3-dimensional randomly
weighted tensor acting as a generic encoder network to provide a
nonlinear embedding of latent relationships among real-valued vectors.
We show that a trained decoder network in RWTNs can effectively capture
the likelihood of \emph{part-of} relationships at a level of performance
exceeding that of traditional LTNs even if far fewer parameters have to
Thus, even though it is untrained, the randomly drawn NTN is shown to
have great relational expressiveness and acts as a general-purpose
feature extractor the same way a randomly drawn recurrent reservoir in
RC generates features for time-series data. To the best of our
knowledge, this is the first research to integrate both RC and SRL
approaches for reasoning under uncertainty and learning in the presence
of data and rich knowledge. Furthermore, because the NTN
weights do not depend upon the data, the single NTN network can be
shared among several decoder networks each trained for a different
classifier, giving RWTNs an economy of spatial scale in applications
where several classifiers need to be used simultaneously.


\section{Related Work}
\label{sec:related}
RWTNs are greatly influenced by LTNs and can be viewed as a
performance-based refactoring of the neural network architecture. The
model theory underlying LTNs was first proposed by
\citet{guha2015towards}; it represents logical terms and predicates
using points/vectors in a $n$-dimensional real space and computes the
truth value of atomic formulas by comparing the projections of the
real-valued vector. By extending the theory and generalizing
NTNs~\citep{socher2013reasoning}, LTNs~\citep{serafini2016logic}~(and
thus also RWTNs) provide more general interpretation of predicate
symbols in first-order logic.

Another neural-network approach for logical representation comes
from~(Hybrid) Markov Logic Networks~(MLNs)~\citep{richardson2006markov,
wang2008hybrid, nath2015learning}. In MLNs, the number of models that
satisfy a formula determines the truth value of the formula. That is,
the more models there are, the higher the degree of truth. Hybrid MLNs
introduce a dependency from real features associated to constants, which
is given and not learned. In our model, instead, the truth value of a
complex formula is determined by~(fuzzy) logical reasoning, and the
relations between the features of different objects is learned through
error minimization.



\section{Preliminaries}
\label{sec:prelim}

\subsection{Reservoir Computing}
\label{sec:RC}

Reservoir Computing~(RC) is a less conventional method for using
Recurrent Neural Networks that has been widely used in applications such
as time-series forecasting~\citep{deihimi2012application,
bianchi2015prediction, bianchi2015short}, process
modelling~\citep{rodan2017bidirectional},
and classification of multivariant time
series~\citep{bianchi2018reservoir}. RC models conceptually divide
time-series processing into two components:~(i)~representation of
temporal structure in the input stream through a non-adaptable dynamic
\emph{reservoir}~(generated through the feedback-driven dynamics of a
randomly drawn RNN), and~(ii)~an easy-to-adapt \emph{readout} from the
reservoir. The feedbacks within the reservoir network provide internal
dynamic state variables allowing the network to re-shape and extend the
duration of short patterns in time, effectively allowing the readout
network to have access to ``echos'' of past versions of the input data.
Consequently, RC techniques were originally introduced to the machine
learning community under the name Echo State
Networks~(ESNs)~\citep{jaeger2001echo}; in this paper, we use the two
terms interchangeably.


The simplest formulation of the recurrent mapping from input to the
internal state of the ESN is:
\begin{equation}
    \label{esn_eq1}
    \textbf{h}(t) = f(\textbf{W}_{in}\textbf{x}(t)+\textbf{W}_{r}\textbf{h}(t-1))
\end{equation}
where $\textbf{h}(t)$ is the internal state of the ESN at time $t$,
which depends upon its previous state $\textbf{h}(t-1)$ and the current
input $\textbf{x}(t)$ by way of $f(\cdot)$, a nonlinear activation
function~(usually a sigmoid or hyperbolic tangent), and the encoder
parameters $\{\textbf{W}_{in}, \textbf{W}_{r}\}$ that are randomly
generated and left untrained~(or implemented using a prefixed
topology~\citep{rodan2010minimum}).
The favorable capabilities of the reservoir primarily depend on three
factors:~(i) a large number of processing units in the recurrent
layer,~(ii) random connectivity of the recurrent layer, and~(iii) a
spectral radius\footnote[1]{The magnitude of the largest eigenvalue,
which can be a rough measure of the global scaling of the weights in the
case of an even eigenvalue spread.} of the connection weights matrix
$\textbf{W}_r$, set to bring the system to the edge of
stability~\citep{bianchi2016investigating}. Therefore, rather than
training the internal weight matrices, the behavior of the reservoir can
be controlled by simply modifying: the spectral radius~$\rho$, the
percentage of non-zero connections~$\beta$, and the number of hidden
units~$R$. Another important hyperparameter is the scaling $\omega$ of
the values in $\textbf{W}_{in}$, which controls the degree of
nonlinearity in the processing units and, jointly with $\rho$, can
change the internal dynamics from a chaotic system to a contractive one.
Finally, for the purpose of regularization, a Gaussian noise with
standard deviation $\xi$ can be added to the state update
function~(Eq.~(\ref{esn_eq1})) as an argument~\citep{jaeger2001echo}.

From the sequence of the ESN states generated over time, described by
the matrix $\textbf{H} = [\textbf{h}(1), \dots, \textbf{h}(T)]^{T}$, it
is possible to define an encoding~(representation) $r(\textbf{H}) =
\textbf{r}_{\textbf{x}}$ of the input sequence $\textbf{x}$. Such a
state becomes a vector representation with a fixed-size and can be
processed by regular machine learning algorithms. Specifically, the
decoder maps the input representation $\textbf{r}_{\textbf{x}}$ into the
output space containing all class labels $\textbf{y}$ in a
classification task:
\begin{equation}
    \label{esn_eq2}
    \textbf{y}=g(\textbf{r}_{\textbf{X}}) = \textbf{V}_{o}\textbf{r}_{\textbf{x}} + \textbf{v}_{o}
\end{equation}
The decoder parameters $\{\textbf{V}_{o}, \textbf{v}_{o}\}$ can be
learned by minimizing a ridge regression loss function
\begin{equation}
    \label{esn_eq3}
    \{\textbf{V}_{o}, \textbf{v}_{o}\}^* = \operatorname*{\arg\min}_{\{\textbf{V}_{o}, \textbf{v}_{o}\}} \frac{1}{2} || \textbf{V}_{o} \textbf{r}_{\textbf{x}} + \textbf{v}_{o} - \textbf{y}||^2 + \lambda||\textbf{V}_{o}||^2, 
\end{equation}
which admits a closed-form solution~\citep{scardapane2017randomness}.

\subsection{Logic Tensor Networks}
\label{sec:LTNs}

Logic Tensor Networks integrate learning based on
NTNs~\citep{socher2013reasoning} with reasoning using first-order,
many-valued logic~\citep{bergmann2008introduction}, all implemented in
TENSORFLOW\textsuperscript{TM}~\citep{serafini2016logic}. This enables a
range of knowledge-based tasks using rich knowledge representation in
First-Order Logic~(FOL) to be combined with efficient data-driven
machine learning.
\subsubsection{First-Order Logic:} A FOL language $\mathcal{L}$ and its
signature consists of three disjoint sets---$\mathcal{C}, \mathcal{F}$
and $\mathcal{P}$---denoting constants, functions and predicate symbols,
respectively. For any function or predicate symbol $s$, $\alpha(s)$ can
be described as its \emph{arity.} Logical formulas in $\mathcal{L}$
enable the description of relational knowledge.
The objects being reasoned over with FOL are mapped to an interpretation
domain, which is a subset of $\mathbb{R}^n$ so that every object is
associated with an $n$-dimensional vector of real numbers. Intuitively,
this $n$-tuple indicates $n$ numerical features of an object. Thus,
functions are interpreted as real-valued functions, and predicates are
interpreted as fuzzy relations on real vectors. With this numerical
background, we can now define the numerical \emph{grounding} of FOL with
the following semantics; this grounding is necessary for NTNs to reason
over logical statements.

Let $n \in \mathbb{N}$. An $n$-grounding, or simply grounding,
$\mathcal{G}$ for a FOL $\mathcal{L}$ is a function defined on the
signature of $\mathcal{L}$ satisfying the following conditions:
\begin{itemize}
    \item $\mathcal{G}(c) \in \mathbb{R}^n$ for every constant symbol $c
        \in \mathcal{C};$
    \item $\mathcal{G}(f) \in \mathbb{R}^{n \cdot \alpha(f)} \to
        \mathbb{R}^n$ for function symbol $f \in \mathcal{F};$
    \item $\mathcal{G}(P) \in \mathbb{R}^{n \cdot \alpha(f)} \to [0,1]$
        for predicate sym.~$P \in \mathcal{P};$
\end{itemize}
Given a grounding $\mathcal{G}$, the semantics of closed terms and
atomic formulas is defined as follows:
\begin{gather*}
    \mathcal{G}(f(t_1, \dots, t_m)) = \mathcal{G}(f)(\mathcal{G}(t_1),
        \dots, \mathcal{G}(t_m)) \\
    \mathcal{G}(P(t_1, \dots, t_m)) = \mathcal{G}(P)(\mathcal{G}(t_1),
        \dots, \mathcal{G}(t_m))
\end{gather*}
According to fuzzy logic such as the Lukasiewicz
$t$-norm~\citep{bergmann2008introduction}, the semantics for connectives
is defined as follows:
\begin{gather*}
    \mathcal{G}(\neg \phi) = 1-\mathcal{G}(\phi) \\
    \mathcal{G}(\phi \land \psi) = \max (0, \mathcal{G}(\phi) + \mathcal{G}(\psi) - 1) \\
    \mathcal{G}(\phi \lor \psi) = \min (1, \mathcal{G}(\phi) + \mathcal{G}(\psi)) \\
    \mathcal{G}(\phi \rightarrow \psi) = \min (1, 1 - \mathcal{G}(\phi) + \mathcal{G}(\psi))
\end{gather*}

\subsubsection{Learning as Best Satisfiability:} A partial grounding
$\hat{\mathcal{G}}$ can be defined on a subset of the signature of
$\mathcal{L}$. A grounding $\mathcal{G}$ is said to be a completion of
$\hat{\mathcal{G}}$ if $\mathcal{G}$ is a grounding for $\mathcal{L}$
and coincides with $\hat{\mathcal{G}}$ on the symbols where
$\hat{\mathcal{G}}$ is defined.

Let \emph{GT} be a grounded theory which is a pair $\langle \mathcal{K},
\hat{\mathcal{G}}\rangle$ with a set $\mathcal{K}$ of closed formulas
and a partial grounding $\hat{\mathcal{G}}$. A grounding $\mathcal{G}$
satisfies a \emph{GT} $\langle \mathcal{K}, \hat{\mathcal{G}}\rangle$ if
$\mathcal{G}$ completes $\hat{\mathcal{G}}$ and $\mathcal{G}(\phi)=1$
for all $\phi \in \mathcal{K}$. A \emph{GT} $\langle \mathcal{K},
\hat{\mathcal{G}}\rangle$ is satisfiable if there exists a grounding
$\mathcal{G}$ that satisfies $\langle \mathcal{K},
\hat{\mathcal{G}}\rangle$. In other words, deciding the satisfiability
of $\langle \mathcal{K}, \hat{\mathcal{G}}\rangle$ amounts to searching
for a grounding $\mathcal{G}$ such that all the formulas of
$\mathcal{K}$ are mapped to 1.  Differently from classical
satisfiability, when a \emph{GT} is not satisfiable, we are interested
in the best possible satisfaction that we can reach with a grounding.
This is defined as follows. Let $\langle \mathcal{K},
\hat{\mathcal{G}}\rangle$ be a grounded theory. We define the best
satisfiability problem as the problem of finding a grounding
$\mathcal{G}^*$ that maximizes the truth values of the conjunction of
all clauses $cl \in \mathcal{K}$, i.e., $\mathcal{G}^* =
\arg\max_{\hat{\mathcal{G}} \subseteq \mathcal{G} \in \mathbb{G}}
\mathcal{G}(\bigwedge_{cl \in \mathcal{K}} cl)$.

\subsubsection{Logical Grounding and NTNs:} Grounding $\mathcal{G}^*$
captures the implicit correlation between quantitative features of
objects and their categorical/relational properties. We consider
groundings of the following form.

Function symbols are grounded to linear transformations. If $f$ is a
$m$-ary function symbol, then  $\mathcal{G}(f)$ is of the form:
\begin{equation*}
    \mathcal{G}(f)(\textbf{\texttt{v}}) = M_f\textbf{\texttt{v}} + N_f
\end{equation*}
where $\textbf{\texttt{v}} = \langle \textbf{\texttt{v}}_1^\top, \dots,
\textbf{\texttt{v}}_m^\top  \rangle^\top$ is the $mn$-ary vector
obtained by concatenating each $\textbf{\texttt{v}}_i$. The parameters
for $\mathcal{G}(f)$ are the $n \times mn$ real matrix $M_f$ and the
$n$-vector $N_f$. The grounding of an $m$-ary predicate $P$, namely
$\mathcal{G}(P)$, is defined as a generalization of the
NTN~\citep{socher2013reasoning}, as a function from $\mathbb{R}^{mn}$ to
$[0,1]$, as follows:
\begin{equation}
    \label{ltn_eq_predicate}
    \mathcal{G}(P)(\textbf{\texttt{v}}) = \sigma (u_{P}^{\top} \texttt{f} (\textbf{\texttt{v}}^{\top} W_{P}^{[1:k]} \textbf{\texttt{v}} + V_{P} \textbf{\texttt{v}} + b_P))
\end{equation}
where $\sigma$ is the sigmoid function and $\texttt{f}$ is the
hyperbolic tangent ($\tanh$). The parameters for $P$ are:
$W_{P}^{[1:k]}$, a 3-D tensor in $\mathbb{R}^{k \times mn \times mn},
V_{P} \in \mathbb{R}^{k \times mn}, b_P \in \mathbb{R}^{k}$ and $u_P \in
\mathbb{R}^k$.

\section{Randomly Weighted Tensor Networks}
\label{sec:RWTNs}

Here, we introduce the mathematical and structural definitions of
Randomly Weighted Tensor Networks~(RWTNs). By combining a randomly
drawn, untrained tensor into an NTN encoder network with a trained
decoder network, our model has fewer parameters to learn and can also
achieve greater expressive capability for extracting relational
knowledge as an LTN trained for the same task.

\subsection{The Definition of RWTNs}
\begin{figure}[t!]
    \vskip 0.1in
    \begin{center}
    \centerline{\includegraphics[width=0.45\textwidth]{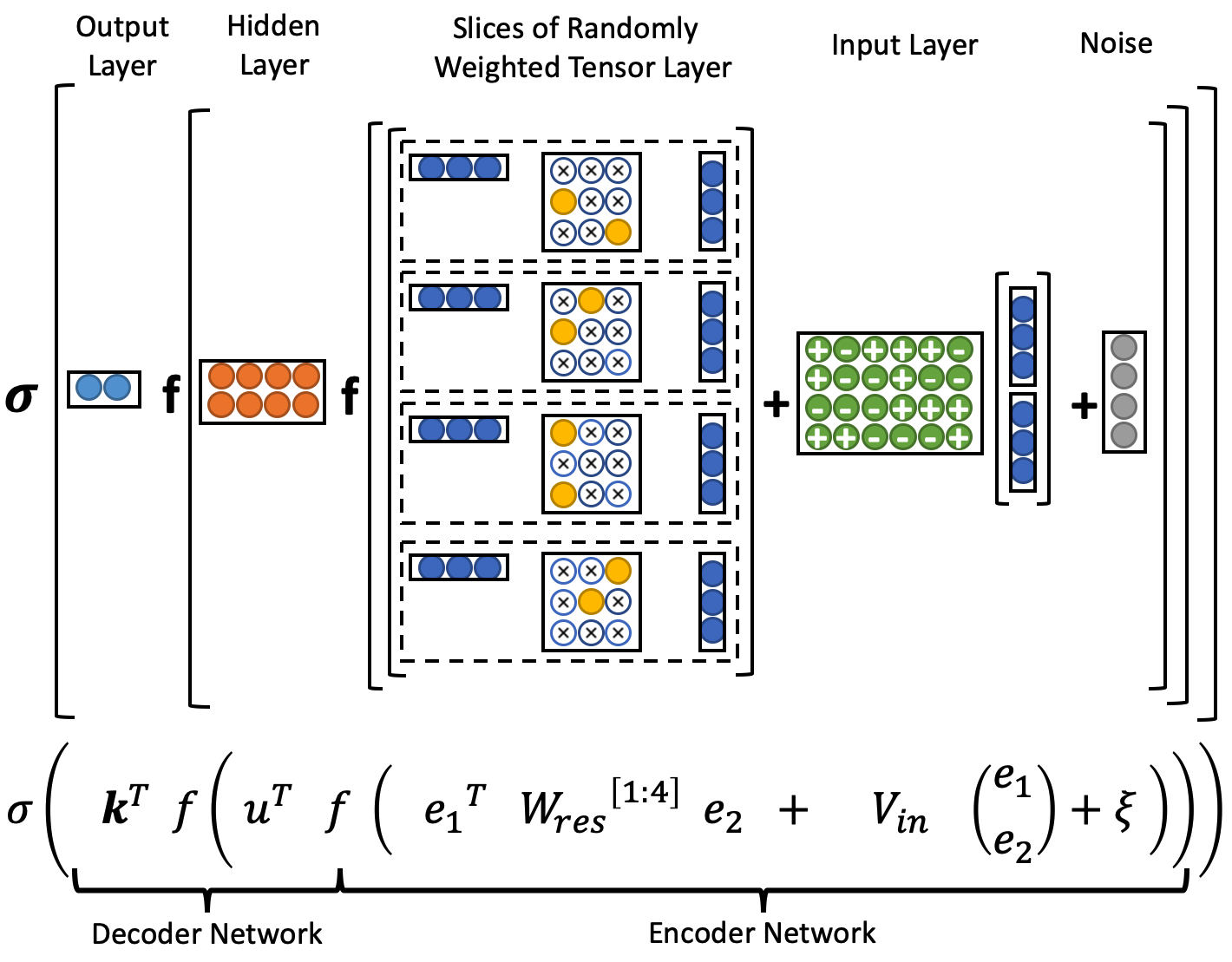}}
    \caption{Visualization of the structure of the Randomly Weighted Tensor Network. Each dashed box represents one slice of the randomly weighted tensor $W_{res}^{[1:R]}$, in this case there are $R=4$ slices.}
    \label{rtn-structure}
    \end{center}
    \vskip -0.1in
\end{figure}

RWTNs can be defined as a function from $\mathbb{R}^{mn}$ to $[0,1]$:
\begin{equation}
\label{rtn_eq}
    \mathcal{G}_{rwtn}(P)(\textbf{\texttt{v}}) = \sigma (\textbf{k}^\top \texttt{f}(u^\top \texttt{f}(\textbf{\texttt{v}}^\top W_{res}^{[1:R]} \textbf{\texttt{v}} + V_{in}\textbf{\texttt{v}} + \xi)))
\end{equation}
where $\sigma$ is the sigmoid function and $\texttt{f}$ is the
hyberbolic tangent ($\tanh$) function. The parameters of the RWTN
encoder include: $W_{res}^{[1:R]} \in \mathbb{R}^{mn \times mn \times
R}$ (a 3-dimensional randomly weighted tensor), $V_{in} \in
\mathbb{R}^{R \times mn}$ (randomly drawn input-layer weights), and
$\xi$ (Gaussian noise). The parameters of the RWTN decoder are thus $u
\in \mathbb{R}^{R \times t}$ and $\textbf{k} \in \mathbb{R}^t$, which
are the standard weights for a single hidden layer neural network where
$t$ is the number of neurons in a hidden layer. Fig.~\ref{rtn-structure}
shows a sample visualization of the structure of our model.

In the depicted case, $e_1, e_2 \in \mathbb{R}^d$ are vector
representations~(or features) of two entities for which the RWTN
expresses some level relationship between. Each slice of the tensor
$W_{res}^{[1:R]}$ can be viewed as being responsible for representing
one kind of relationship between the two entities. In principle, the
network could be trained to explicitly represent certain relationships.
However, this tensor is randomly weighted in RWTN to span a wide range
of potential relationships that are left to the later decoder to mix to
represent the desired relationships from data. There are the following
characteristics of our model:
\begin{itemize}
    \item Non-adaptability of the parameters in the encoder network,
        inspired by the insights of Reservoir Computing~(RC): the tensor
        $W_{res}^{[1:R]}$ are selected to have a greater number of
        units, random sparsity, and a certain spectral radius. Also, the
        input weights $V_{in}$ are generated randomly from a uniform
        distribution over an interval $[-\omega, \omega]$ with random
        sign determined by a random draw from Bernoulli
        distribution~(input-layer weights in Fig.~\ref{rtn-structure}). A
        Gaussian noise $\xi$ is used for the same purpose of the one in
        RC, which is regularization. We intend that by having those
        properties, the randomly weighted tensor and input weights in
        our model can act as a filter that converts the latent
        relationship between objects using a high-dimensional map,
        similar to the operation of an explicit, temporal kernel
        function.
    \item Succinctness in learning process of a decoder network: Using a
        single hidden layer neural network as a decoder enables learning
        the degree of relationship between input entities even though
        far fewer parameters are employed for learning~($\textbf{k}$ and
        $u$) compared to conventional neural tensor networks~(hidden
        output layers in Fig.~\ref{rtn-structure}).
\end{itemize}




\subsection{RWTNs with Weight Sharing}

A unique feature of RWTNs is that because the weights of the randomized
NTN encoder are independent of the training data, the encoder network
can be shared among decoder networks trained for different classifiers.
We refer to this property as \emph{weight sharing}.
\begin{figure}[t!]
\vskip 0.1in
\begin{center}
\centerline{\includegraphics[width=\columnwidth]{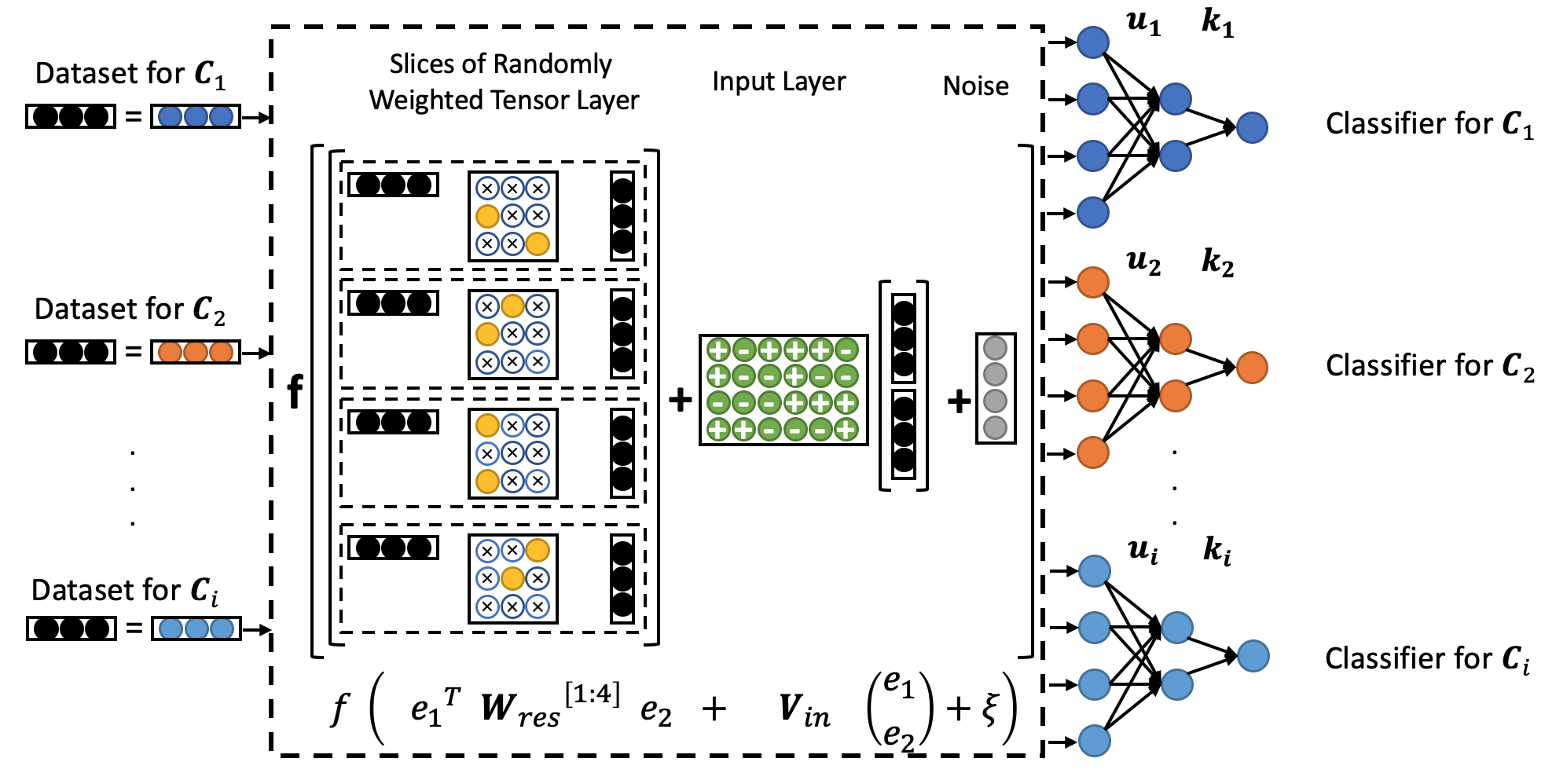}}
\caption{Visualization of the structure of the Randomly Weighted Tensor
Network with weight sharing. In the case of learning each classifier
from the class $\mathcal{C}_{1}$ to the class $\mathcal{C}_{i}$, RWTNs
allow us to use the same encoder network to extract features from each
data from the class $\mathcal{C}_{1}$ to the class $\mathcal{C}_{i}$.}
\label{rtn-ws-structure}
\end{center}
\vskip -0.1in
\end{figure}
Fig.~\ref{rtn-ws-structure} shows a visualization of the structure of
our model applied with weight sharing to the learning of $i$ different
classifiers. The large dashed box surrounds a single encoder network
that serves as a common feature extractor for all classifiers.
Instead of generating the encoder network for each classifier (as in a
conventional LTN), each classifier uses the same encoder network, and
training only requires learning the weights of that classifier's
relatively simple decoder network. This approach increases reusability
and cost efficiency in a way that goes beyond what is possible with LTNs
which must train all encoder and decoder networks separately for each
classifier.

\section{Experimental Evaluation}
\label{sec:experiment}

To evaluate the performance of our proposed RWTNs over LTNs, we employ
both for Semantic Image Interpretation~(SII) tasks, which extract
structured semantic descriptions from images. Very few SRL applications
have been applied to SII tasks because of the high complexity involved
with image learning. \citet{donadello2017logic} define two main tasks of
SII as:~(i) the classification of bounding boxes, and~(ii) the detection
of the \emph{part-of} relation between any two bounding boxes. They
demonstrated that LTNs can successfully improve the performance of
solely data-driven approaches, including the state-of-the art Fast
Region-based Convolutional Neural Networks~(Fast
R-CNN)~\citep{girshick2015fast}.
Our experiments are conducted by comparing the performance of two tasks
of SII between RWTNs and LTNs. These tasks are well defined in
first-order logic, and the codes implemented in
TENSORFLOW\textsuperscript{TM} have been provided and thus can be easily
used to compare the performance of LTNs with RWTNs.

\subsection{Methods}

We provide details of our experimental comparison of RTWNs and LTNs. In
this section, we introduce (i) how to formalize our two focal SII tasks
in FOL grounded in RWTNs and LTNs, (ii) the data set used in the test,
and (iii) the RWTN and LTN hyperparameters used in the test.

\subsubsection{Formalizing SII in First-Order Logic:}
\label{sec:formalizingSII}

A signature $\Sigma_{\textsc{SII}} = \langle \mathcal{C, F, P} \rangle$
is defined where $C = \bigcup_{p \in Pics} b(p)$ is the set of
identifiers for all the bounding boxes in all the images, $\mathcal{F} =
\emptyset$, and $\mathcal{P} = \{\mathcal{P}_1, \mathcal{P}_2\}$, where
$\mathcal{P}_1$ is a set of unary predicates, one for each object type
(e.g., $\mathcal{P}_1 = \{ \texttt{Dog}, \texttt{Tail}, \dots\}$), and
$\mathcal{P}_2$ is a set of binary predicates representing relations
between objects. Because our experiments focus on the \emph{part-of}
relation, $\mathcal{P}_2 = \{\texttt{partOf}\}$. The FOL formulas based
on this signature can specify:~(i) simple facts; 
the fact that $b$ contains either a cat or a dog
, or~(ii) general rules.

We define the grounding for $\Sigma_{\textsc{SII}}$ such that each
constant $b$, indicating a bounding box, is associated with geometric
features describing the position and the dimension of the bounding box
and semantic features indicating the classification score returned by
the bounding box detector for each class. For example, for each bounding
box $b \in \mathcal{C}, C_i \in \mathcal{P}_i, \mathcal{G}(b)$ is the
$\mathbb{R}^{4+|\mathcal{P}_1|}$ vector:
\begin{multline*} \langle class(C_1, b), \dots,
class(C_{|\mathcal{P}_1|}, b), \dots\\ x_0(b), y_0(b), x_1(b), y_1(b)
\rangle \end{multline*}
where the last four features are the coordinates of the top-left and
bottom-right corners of $b$, and $class(C_i, b) \in [0,1]$ is the
classification score of the bounding box detector for $b$.
For each class $C_i \in \mathcal{P}_1$, define the grounding:
\begin{equation}
\label{object_classification_eq}
    \mathcal{G}(C_i)(\textbf{x}) =
    \begin{cases}
    1, & \text{if } i = \arg\max_{1 \leq l \leq |\mathcal{P}_1|} x_l \\
    0 & \text{otherwise}
    \end{cases}
\end{equation}
where $\textbf{x} = \langle x_1, \dots, x_{4+|\mathcal{P}_1|} \rangle$
is the vector corresponding to the grounding of a bounding box.

$\mathcal{G}(\texttt{partOf}(b, b'))$ can be defined as:
\begin{equation}
\label{partof_detection_eq}
    \begin{cases}
    1, & \text{if } ir(b, b') \cdot \max_{ij=1}^{|\mathcal{P}_1|} (w_{ij} \cdot x_i \cdot x'_j) \geq th_{ir} \\
    0 & \text{otherwise}
    \end{cases}
\end{equation}
for some threshold $th_{ir}$~(usually, $th_{ir} > 0.5$) where $ir(b, b')
= \frac{area(b \cap b')}{area(b)}$ and $w_{ij} = 1$ if $C_i$ is a part
of $C_j$ ($w_{ij}=0$ otherwise). Given the above grounding, we can
compute the grounding of any atomic formula thus expressing the degree
of truth of the formula.

\subsubsection{Defining the Grounded Theories for RWTNs and LTNs:}
\label{sec:groundingRWTNs}

A suitable ground theory \emph{GT} can be built for SII. Let $Pics^t
\subseteq Pics$ be a set of bounding boxes of images correctly labelled
with the classes that they belong to, and let each pair of bounding
boxes be correctly labelled with the \emph{part-of} relation. Then,
$Pics^t$ can be considered as a training set and a grounded theory
$\mathcal{T}_{\text{LTN}}$ can be constructed as follows:
$\mathcal{T}_{\text{LTN}} = \langle \mathcal{K}, \hat{\mathcal{G}}
\rangle$, where:
\begin{itemize}
    \item $\mathcal{K}$ contains: (i) the set of closed literals
        $C_i(b)$ and $\texttt{partOf}(b, b')$ for every bounding box $b$
        labelled with $C_i$ and for every pair of bounding boxes
        $\langle b, b' \rangle$ connected by the \texttt{partOf}
        relation, and~(ii) the set of the mereological constraints for
        the \emph{part-of} relation, including asymmetric constraints,
        lists of several parts of an object, or restrictions that whole
        objects cannot be part of other objects and every part object
        cannot be divided further into parts. 
    \item The partial grounding $\hat{\mathcal{G}}$ is defined on all
        bounding boxes of all the images in $Pics$ where both
        $class(C_i,b)$ and the bounding box coordinates are computed by
        the Fast R-CNN object detector. $\hat{\mathcal{G}}$ is not
        defined for the predicate symbols in $\mathcal{P}$ and is to be
        learned.
\end{itemize}

A grounded theory $\mathcal{T}_{\text{RWTN}}$ is only slightly
different. $\mathcal{T}_{\text{RWTN}} = \langle \mathcal{K},
\hat{\mathcal{G}}_{rwtn} \rangle$ where a partial grounding
$\hat{\mathcal{G}}_{rwtn}$ can be described for predicates using
eq.(\ref{rtn_eq}). Thus, we can easily compare the performance between
$\hat{\mathcal{G}}_{rwtn}$ and $\hat{\mathcal{G}}$.

\subsubsection{Datasets:}
\label{sec:datasets}

The \textsc{PASCAL-Part}-dataset~\citep{chen2014detect} and
ontologies~(\textsc{WordNet}) are chosen for the \emph{part-Of}
relation. The \textsc{PASCAL-Part}-dataset contains 10103 images with
bounding boxes. They are annotated with object-types and the part-of
relation defined between pairs of bounding boxes. There are three main
groups in labels---animals, vehicles, and indoor objects---with their
corresponding parts and ``part-of'' label. There are 59 labels~(20
labels for whole objects and 39 labels for parts).
The images were then split into a training set with 80\% of the images
and a test set with 20\% of the images, maintaining the same proportion
of the number of bounding boxes for each label. Given a set of bounding
boxes detected by an object detector~(Fast-RCNN), the task of object
classification is to assign to each bounding box an object type. The
task of \emph{part-Of} detection is to decide, given two bounding boxes,
if the object contained in the first is a part of the object contained
in the second.

\subsubsection{Hyperparameter Setting:}
\label{sec:hyperparameters}

To compare the performance between RWTNs and LTNs, we train two models
separately.
\begin{itemize}
    \item For RWTN, the spectral radius $\rho$ is set to 0.6, the
        connection sparsity $\beta$ is 0.25. The size of the reservoir
        $R$ is 200. The input scaling $\omega$ is 0.5. The noise level
        $\xi$ is 0.01. The number of hidden units for a readout $t$ is
        20.
    \item For LTN, we configure the experimental environment
        following~\citet{donadello2017logic}. The LTNs were configured
        with a tensor of $k=6$ layers.
\end{itemize}
Both models make use of a regularization parameter $\lambda=10^{-10}$,
Lukasiewicz’s $t$-norm~($\mu(a, b) = \operatorname{max}(0, a + b - 1)$),
and the harmonic mean as an aggregation operator. We ran 1000 training
epochs of the RMSProp learning algorithm available in
TENSORFLOW\textsuperscript{TM} for each model.

\subsection{Results}
\label{sec:results}

Our experiments mainly focus on the comparison of the performance
between our model and LTN, but figures also include the results with the
Fast-RCNN~\citep{girshick2015fast} at type
classification~(Eq.~(\ref{object_classification_eq})) and the inclusion
ratio~$ir$ baseline~(Eq.~(\ref{partof_detection_eq})) at the
\emph{part-Of} detection task. If $ir$ is greater than a given threshold
$th$~(in our experiments, $th = 0.7$), then the bounding boxes are said
to be in the \texttt{partOf} relation. Every bounding box $b$ is
classified into $C \in \mathcal{P}_1$ if $\mathcal{G}(C(b)) > th$.

\begin{figure*}[!h!t]
\vskip 0.1in
\centering
  \subfigure[RWTNs sometimes outperform LTNs on object type classification, achieving an Area Under the Curve~(AUC) of 0.777 in comparison with 0.766.]{\includegraphics[scale=.3]{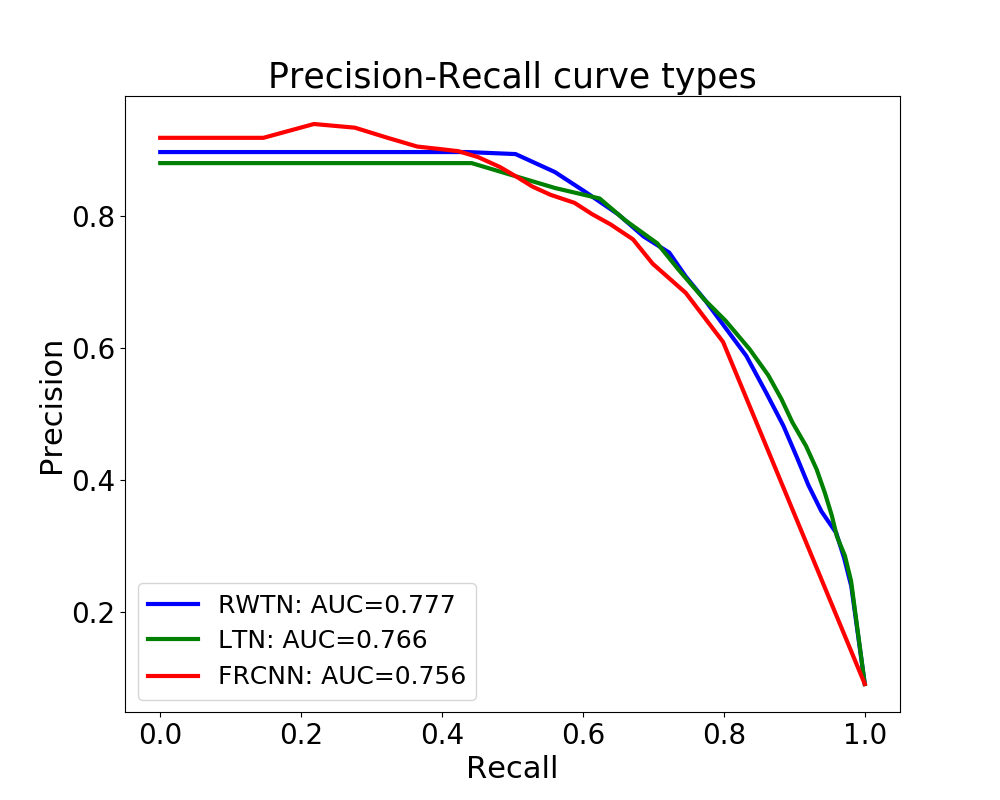}\label{sii-indoor-result01}}
  \quad
  \subfigure[RWTNs mostly show better performance than LTNs in the detection of \emph{part-of} relations, achieving AUC of 0.661 in comparison with 0.620.]{\includegraphics[scale=.3]{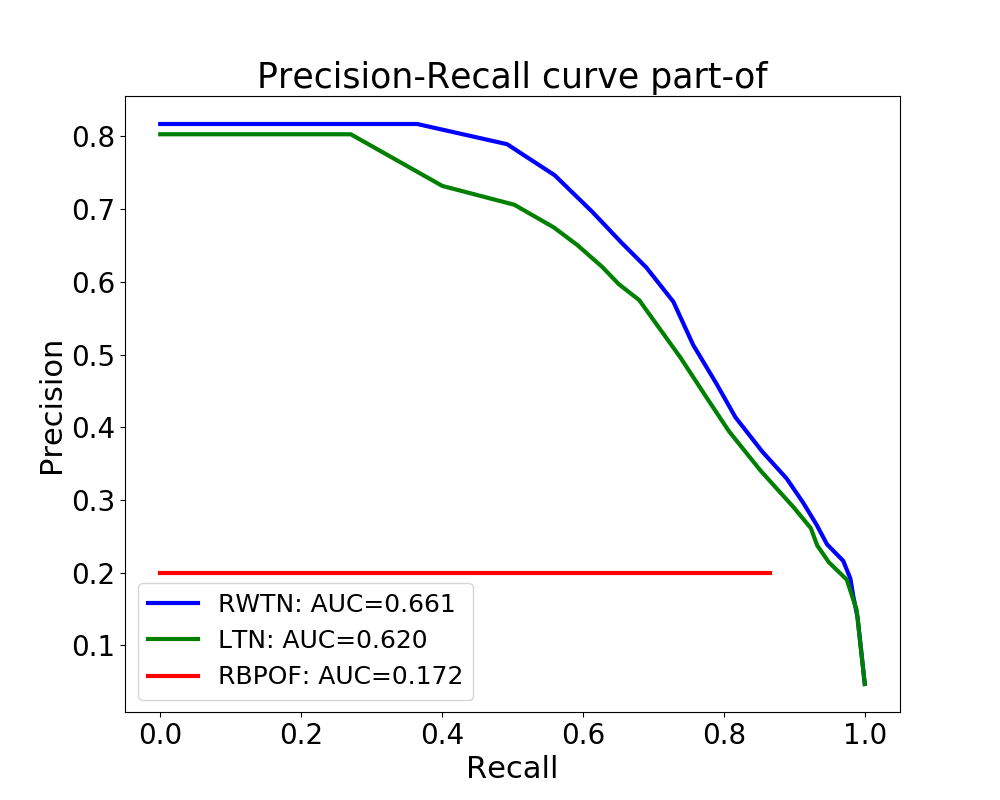}\label{sii-indoor-result02}}
  \caption{Precision--recall curves for indoor objects type classification and the \texttt{partOf} relation between objects.}
  \label{sii-indoor-results}
 \vskip -0.1in
\end{figure*}
Results for indoor objects are shown in Fig.~\ref{sii-indoor-results}
where AUC is the area under the precision--recall curve. The results
show that, for the \emph{part-Of} relation and object types
classification, RWTNs achieve better performance than LTNs.
%
However, there are some variance in the results because of the
stochastic nature of the experiments. Consequently, we carried out five
such experiments for each task, for which the sample averages and 95\%
confidence intervals are shown in Table~\ref{table_performance}. These
results confirm that our model can achieve similar performance as LTNs
for object-task classification and superior performance for detection of
\emph{part-of} relations.

In Table~\ref{table_performance}, we only included AUC numbers for RWTNs
with weight sharing (third column) for object-type classification
because \emph{part-of} relations only require a single classifier. The
performance of RWTNs with weight sharing for the object-type
classification task (which requires 11 classifiers for indoor objects,
23 for vehicles, and 26 for animals) shows only a marginal gap in
performance compared to other models, which demonstrates the
effectiveness and efficiency of the approach of using a shared,
randomized NTN in RWTNs with weight sharing.
%
\begin{table}[b!]
\caption{AUC of T1~(object type classification) and T2~(detection of \emph{part-of} relation) for LTN, RWTN, and RWTN with weight sharing across label groups}
\label{table_performance}
\vskip 0.1in
\begin{center}
\begin{small}
\begin{sc}
\begin{tabular}{lcccc}
\toprule
Label & Tasks & LTN & RWTN & RWTN w/ W.S \\
\midrule
\multirow{2}{1em}{Indoor} & T1 & \textbf{.77 $\pm$ .027} & \textbf{.77 $\pm$ .012} & .76 $\pm$ .0068 \\
& T2 &  .64 $\pm$ .060& \textbf{.66 $\pm$ .049} & - \\
\multirow{2}{1em}{Vehicle} & T1 & \textbf{.73 $\pm$ .017} & .71 $\pm$ .030 & .70 $\pm$ .014 \\
& T2 &  .53 $\pm$ .065& \textbf{.58 $\pm$ .037} & - \\
\multirow{2}{1em}{Animal} & T1 & \textbf{.69 $\pm$ .028} & \textbf{.69 $\pm$ .024} & .68 $\pm$ .016 \\
& T2 &  .60 $\pm$ .092& \textbf{.64 $\pm$ .070} & - \\
\bottomrule
\end{tabular}
\end{sc}
\end{small}
\end{center}
\vskip -0.1in
\end{table}

\subsection{Relative Complexity of RWTNs and LTNs}
\label{sec:discussion}

To better appreciate the relative performance of RWTNs and LTNs, we can
compare the number of parameters to learn for grounding a unary
predicate for each model. Let $n$ be the number of features of an
input~($n = 64$) for both RWTNs and LTNs. As shown in
Eq.~(\ref{ltn_eq_predicate}), the parameters to learn in LTNs are $\{
u_{P} \in \mathbb{R}^k, W_{P}^{[1:k]} \in \mathbb{R}^{n \times n \times
k}, V_{P} \in \mathbb{R}^{k \times n}, b_P \in \mathbb{R}^k \}$, where
$k = 6$ following the configuration of the LTNs. Thus, the number of
parameters in LTNs is $(n^2 + n + 2) \cdot k = (64^2 + 64 + 2) \cdot 6 =
24972$. On the other hand, in Eq.~(\ref{rtn_eq}), the learnable
parameters in RWTNs are only $\{ \textbf{k} \in \mathbb{R}^t, u \in
\mathbb{R}^{R \times t} \}$, where $R = 200$ and $t = 20$ following the
configuration of the RWTNs. Therefore, the number of learnable
parameters in RWTNs is $(R + 1) \cdot t = 201 \cdot 20 = 4020$. The fact
that the number of parameters to learn in RWTNs~($4020$) is
significantly smaller compared to LTNs~($24972$) shows that
non-adaptable parameters in RWTNs can have significant power to
represent the latent relationship among objects so that the model can
efficiently extract relational knowledge even though using fewer
parameters.
Furthermore, the number of the parameters of LTNs heavily depends on the
number of features, whereas RWTNs are independent of the number of
features. In principle, this could allow the learning process in our
model to be accelerated if the feature representation from the encoder
model is pre-processed and stored.

\subsection{Space Complexity of RWTNs with Weight Sharing}
Weight sharing is a unique feature of RWTNs, which can greatly reduce
necessary space complexity when multiple classifiers are used
simultaneously. In the depicted case of learning $i$ classifiers in
Fig.~\ref{rtn-ws-structure},
space complexity for RWTNs is
$((n^2 + n) \cdot R + (R + 1) \cdot t ) \cdot i \approx O(iRn^2)$.
However, with weight sharing, RWTNs can achieve much better space
complexity, which is $(n^2 + n) \cdot R + (R+1) \cdot t \cdot i \approx
O(Rn^2)$
because $i \cdot t < n^2$ for the experiments conducted in the SII task.
This indicates that the number of classifiers can have a negligible
effect on the spatial complexity of RWTNs when weight sharing is used.

\section{Conclusion and Future Work}
\label{sec:conclusion}

In this paper, we introduced Randomly Weighted Tensor Networks, which,
when compared to a conventional neural tensor model, act as a
generalized feature extractor with greater relational expressiveness and
a learning model with relatively simpler structure. We demonstrated how
insights from Reservoir Computing normally reserved for time-series
analysis can be applied to the fields of neural-symbolic computing and
knowledge representation and reasoning for relational learning.

Our work can be advanced in several ways.
We will investigate how other methods from reservoir computing for
exploring efficient reservoir topologies~\citep{ferreira2009genetic,
sun2017deep, wang2019echo} might be generalized to these new application
spaces.
In addition, we shall extend RWTNs to include a recurrent part for
representing dynamic features of time-series data; this approach may
allow for extracting time-varying relational knowledge necessary for
developing a framework for data-driven reasoning over temporal logic.
In addition, ensemble learning may be able to capitalize on blending
across RWTNs with different random realizations.

\section{Acknowledgments}
\label{sec:acknowledgements}

Funding was provided by contract number FA8651-17-F-1013 from the
USAF/Eglin AFB/FL contract number W31P4Q18-C-0054 from DARPA.

\bibliography{main}
\end{document}